\documentclass[runningheads]{llncs}
\usepackage{graphicx}
\usepackage{amsmath}
\usepackage[colorlinks,citecolor=blue,urlcolor=blue]{hyperref}
\usepackage{bbding}
\usepackage{bbm}
\usepackage{cite}
\usepackage{multirow}
\usepackage[misc]{ifsym}
\usepackage{url}
\usepackage{amssymb}
\newcommand{\tabincell}[2]{\begin{tabular}{@{}#1@{}}#2\end{tabular}}
\begin{document}
\title{Transformer Lesion Tracker}
\titlerunning{Transformer Lesion Tracker}
%
\author{Wen Tang\inst{1}, 
Han Kang\inst{1}, 
Haoyue Zhang\inst{1,2}, 
Pengxin Yu\inst{1}, 
Corey W. Arnold\inst{2},
Rongguo Zhang \inst{1}\textsuperscript{(\Letter)} 
}

\authorrunning{Wen Tang et al.}
%
\institute{InferVision Medical Technology Co., Ltd., Beijing, China\\
\email{zrongguo@infervision.com}, \\
Computational Diagnostic Lab, UCLA, Los Angeles, USA}
\maketitle              
\begin{abstract}
Evaluating lesion progression and treatment response via longitudinal lesion tracking plays a critical role in clinical practice. Automated approaches for this task are motivated by prohibitive labor costs and time consumption when lesion matching is done manually. Previous methods typically lack the integration of local and global information. In this work, we propose a transformer-based approach, termed Transformer Lesion Tracker (TLT). Specifically, we design a Cross Attention-based Transformer (CAT) to capture and combine both global and local information to enhance feature extraction. We also develop a Registration-based Anatomical Attention Module (RAAM) to introduce anatomical information to CAT so that it can focus on useful feature knowledge. A Sparse Selection Strategy (SSS) is presented for selecting features and reducing memory footprint in Transformer training. In addition, we use a global regression to further improve model performance. We conduct experiments on a public dataset to show the superiority of our method and find that our model performance has improved the average Euclidean center error by at least 14.3\% (6mm vs. 7mm) compared with the state-of-the-art (SOTA). Code is available at \url{ https://github.com/TangWen920812/TLT}.
\end{abstract}
\section{Introduction}
The ability to accurately locate the location of follow-up lesions and subsequent quantitative assessment, referred to as "lesion tracking," is crucial to a variety of medical applications, in particular, cancer management. In practice, physicians need to spend significant time and effort to precisely match the same lesion across different time points. Thus, its investigation into a fully automated method of lesion tracking or lesion matching is highly desirable.  
However, compared with a large number of studies on lesion segmentation and detection~\cite{shao2019attentive, tang2021m}, there are very few studies on lesion tracking~\cite{cai2021deep, hering2021whole}. In the field of natural images, there is a similar problem called target tracking or object tracking, for which several deep learning-based methods have been proposed~\cite{bolme2010visual, li2019evolution, teed2020raft}. One of the simplest and most straightforward ideas is to apply these existing methods to lesion tracking tasks. However, lesion tracking is different from the aforementioned visual tracking in a number of aspects: (1) Medical imaging data are mostly in 3D format. (2) The lesion size varies at different time points, such as increasing, shrinking, or stabilizing. (3) The appearance of the lesion may change during the follow-up examination while its anatomical location remains unchanged. Thus, an effective lesion tracker should account for the differences in the lesion itself and be able to use anatomical information effectively.
However, existing registration-based trackers~\cite{tan2015new, raju2020co} are not robust for small-sized lesions or heavily deformed lesions due to lack of sensitivity to local details, and Siamese networks~\cite{gomariz2019siamese, li2019evolution} overlook the information around the lesion.
Cai et al.~\cite{cai2021deep} first provided an open-source dataset for lesion tracking and designed Deep Lesion Tracker (DLT), which combines the advantages of both strategies and obtained a baseline on this dataset. Although a large kernel size is extracted in cross correlation layers of DLT to encode the global image context, it is still susceptible to the inductive bias in convolution, leading to deviation in the aggregation of information around the lesion.  

In this work, we leverage Transformer architecture, inspired by TransT~\cite{chen2021transformer}, to replace existing cross correlation, and propose a novel Transformer Lesion Tracking framework (named TLT) using 3D features. To achieve our model, we design a Cross Attention-based Transformer (CAT) to capture global information. To better focus on useful features, we also introduce anatomical priors via the proposed Registration-based Anatomical Attention Module (RAAM) into CAT. Meanwhile, considering the memory cost in training process, we present a Sparse Selection Strategy (SSS) to extract the local effective information from the whole template image as input for CAT. In addition, we use a global regression as output to reduce the effect of insufficient multi-scale information and accelerate convergence. The experimental results show that the proposed method achieves better performance on the open-source dataset compared with the state-of-the-art methods.

\section{Related Work}
\textbf{Registration-based Trackers.} The anatomy presented in a patient's medical images at different time points should be similar in the absence of surgery or similar treatment. Thus, lesion tracker should follow a spatial consistency that the tissue or the structure around a lesion, and the organ in which the lesion is located will not change significantly. Under this assumption, existing registration methods, such as Voxelmorph~\cite{balakrishnan2018unsupervised}, provide solutions for lesion tracking. Since registration algorithms~\cite{marstal2016simpleelastix, heinrich2013mrf} focus on alignment or optimization on global structures, registration-based lesion trackers achieve decent performance for large-sized lesions or relatively stable lesions~\cite{tan2015new, raju2020co}. Still, due to the lack of sensitivity of registration algorithms to image details, these registration-based methods obtain reduced performance when dealing with small-sized lesions or heavily deformed lesions. In this study, we treat image registration as an auxiliary operation, thereby improving model training efficiency as well as performance. Specifically, we select the mask registered via an affine registration method~\cite{marstal2016simpleelastix} as the prior attention and introduce it into the Transformer. The subsequent ablation experiment results demonstrate the effectiveness of this operation. \\
\noindent\textbf{Siamese Networks.} In recent years, Siamese-based methods have been popular in the field of visual object tracking. SiamFC~\cite{bertinetto2016fully}, and its variants such as SiamRPN~\cite{li2018high} and SiamRPN++~\cite{li2019evolution}, are among the representative works. Subsequently, existing studies have demonstrated that lesion tracking could also be done using Siamese-based methods. Gomariz et al.~\cite{gomariz2019siamese} and Liu et al.~\cite{liu2020cascaded} applied 2D Siamese networks for lesion tracking in ultrasound sequences. Whereas Rafael-Palou et al.~\cite{rafael2021re} performed 3D Siamese networks to track lung nodules on CT series. Cai et al.~\cite{cai2021deep} followed SiamRPN++ to use 3D Siamese networks to conduct universal lesion tracking in whole body CT images. These Siamese-based methods mainly consist of two parts: a backbone network for feature extraction and a correlation module to calculate the similarity between the template patch and the searching sub-region. However, such module is susceptible to the inductive bias of convolution operation and fails to fully utilize the global context, leading to local optimum in the optimization process. Thus, we introduce an attention-based Transformer architecture to focus on the key object in the global feature space, while replacing the correlation part.\\
\noindent\textbf{Transformer-based Tracking.} In recent years, Transformer architecture~\cite{vaswani2017attention} has taken over recurrent neural networks in natural language processing~\cite{devlin2018bert, synnaeve2019end}, and has also had an impact on the status of convolutional neural networks in computer vision~\cite{parmar2018image, carion2020end}. More recently, Chen et al.~\cite{chen2021transformer} proposed a target tracking method on natural images with Transformer architecture instead of the cross correlation layers and achieved SOTA results. However, several issues remain to be addressed when applying Transformer to lesion tracking on 3D medical images. Specifically, to reduce memory cost and acquire features of different sized lesion adaptively, we design a sparse selection strategy to extract irregular patches from template feature maps as input to Transformer. To introduce prior anatomical structure information to Transformer, we create a registration-based attention guidance for auxiliary model training.

\section{Method}
\noindent\textbf{Problem Description.} Same as object tracking~\cite{bertinetto2016fully}, lesion tracking aims to find its corresponding position in the searching image $I_{s}$ when given a lesion in the template image $I_{t}$. Similar to~\cite{cai2021deep}, we simplify this task: given a lesion $l$ in $I_{t}$ with its known center $c_{t}$ and radius $r_{t}$, we seek a mapping function $\mathcal{F}$ to predict the center $c_{s}$ of $l$ in $I_{s}$.

\noindent\textbf{Overview.} In this lesion tracking task, we define the baseline CT scan as the template image $I_{t} \in \mathbb{R}^{D_{t0} \times H_{t0} \times W_{t0}}$, and a corresponding follow-up CT scan as the searching image $I_{s} \in \mathbb{R}^{D_{s0} \times H_{s0} \times W_{s0}}$. $D_{t0}$, $H_{t0}$ and $W_{t0}$ represent the depth, height and width of the template image, respectively. And $D_{s0}$, $H_{s0}$ and $W_{s0}$ are similarly defined for searching image. 
The proposed lesion tracking network (TLT) mainly consists of three components, as shown in Fig.~\ref{fig1}. The feature extractor stacks 3D convolution and downsampling layers to efficiently represent the input volumes. The proposed sparse selection strategy is used for memory reducing and efficient feature acquisition. Then, the cross attention-based Transformer (CAT) is used to fuse the features of the searching and the selected template. In the CAT, a mask gained from the registration-based anatomical attention module (RAAM) is inserted to enhance fused features. Finally, the center predictor is responsible for getting the result from the output of Transformer.

\subsection{Feature Extractor and Sparse Selection Strategy} 
\label{sec2.1}
\begin{figure}
\includegraphics[width=\textwidth]{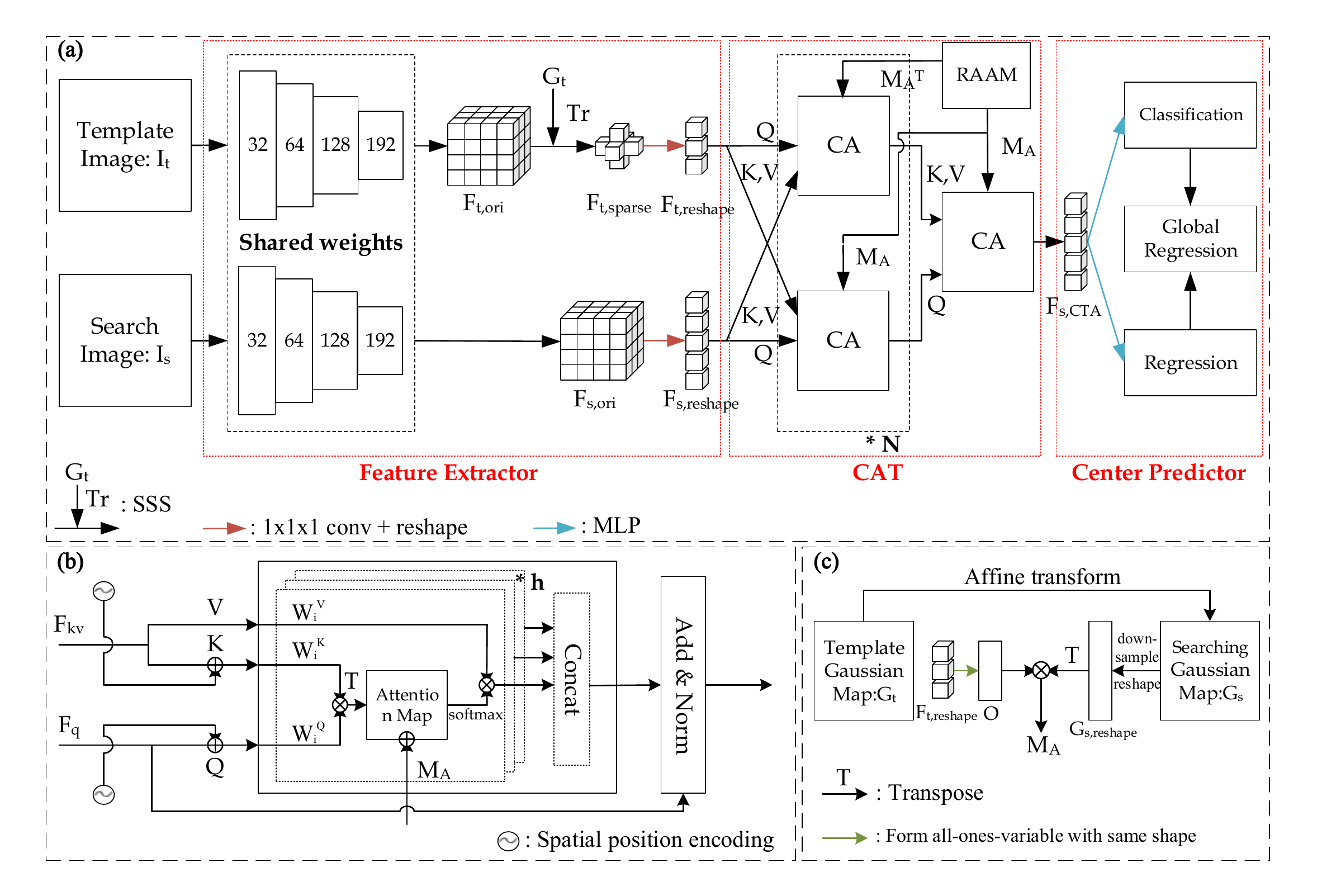}
\caption{(a) Overall structure of the proposed network. (b) Cross attention in CAT. (c) Structure of RAAM} \label{fig1}
\end{figure}

In the proposed network, a modified 3D ResNet18 with shared weights is employed as the feature extractor. Compared with the original one, we remove the last stage of ResNet18, and take outputs of the fourth stage as final outputs. We also adjust the stride of the first convolutional layer from $2\times2\times2$ to $1\times1\times1$ to obtain a larger feature resolution. Considering the parameter redundancy and overfitting in 3D networks, the number of feature channels in each stage is reduced by half or more (see Fig.~\ref{fig1}). 
Putting $I_{t}$ and $I_{s}$ through the learnable feature extractor respectively, their own image features $F_{t,ori} \in \mathbb{R}^{C\times D_{t} \times H_{t} \times W_{t}}$, $F_{s,ori} \in \mathbb{R}^{C\times D_{s} \times H_{s} \times W_{s}}$ are obtained for subsequent processes, where $D_{t}, H_{t}, W_{t} = \frac{D_{t0}}{8}, \frac{H_{t0}}{8}, \frac{W_{t0}}{8}$, $D_{s}, H_{s}, W_{s} = \frac{D_{s0}}{8}, \frac{H_{s0}}{8}, \frac{W_{s0}}{8}$, $C=192$.

As shown in Fig.~\ref{fig1}(a), template-based feature mining via the proposed \textbf{s}parse \textbf{s}election \textbf{s}trategy (SSS) precede the CTA to select features and to reduce memory cost. This is feasible because the location of the lesion in the template input $I_{t}$ is known, and we believe the features $F_{t,ori}$ to have already incorporated local contextual information. The following are the details of the SSS flow. Given a lesion in the template image, we first generate a three-dimensional Gaussian map $G$ based on the known center and radius of the lesion, which is formulated by:
\begin{equation} \label{eq1}
G(c,r) = exp(-\frac{\sum_{i \in (x, y, z)}(i - c^i)^2}{\sum_{i \in (x, y, z)}(2r^i)^2})
\end{equation}
Specifically, for the lesion $l$ in $I_{t}$ with its center $c_{t}$ and radius $r_{t}$ , the generated Gaussian map $G_{t}$ is $G_{t}(c_{t},r_{t})$. Next, we resize $G_{t}$ to the size of $F_{t,ori}$ by trilinear interpolation, and obtain the resized Gaussian map $\widetilde{G}_{t}$. Selecting a threshold $Tr$ and using $\widetilde{G}_{t}$ as a reference mask, we extract valid features $F_{t,sparse}$ from the $F_{t,ori}$ as an input of the Transformer: $F_{t, sparse} = F_{t,ori}(x, y, z | \widetilde{G}_{t}(x, y, z) > Tr)$

\subsection{Cross Attention-based Transformer} \label{sec2.2}
Unlike the similarity-based correlation module used in the previous Siamese-based networks, we design a Cross Attention-based Transformer (CAT) to combine global and local context. Queries $Q$, keys $K$ and values $V$ are encoded from same source in Transformer~\cite{vaswani2017attention}. But in CAT, to grab global context and blend multiple features of different sizes, we adopt cross-attention (CA), in which $K$, $V$ are stemmed from the same input while $Q$ from another. In TLT, we put a CA on each of the template and searching path respectively (see Fig~\ref{fig1}(a)). In CA of the template line, $K$ and $V$ are encoded from the reshaped $F_{s, ori}$ : $F_{s,reshape} \in \mathbb{R}^{D_{s}H_{s}W_{s} \times C}$ and $Q$ is encoded from the reshaped $F_{t, sparse}$ : $F_{t, reshape} \in \mathbb{R}^{L \times C}, L=\text{len}(F_{t, sparse})$. While in CA of the searching line, $K$, $V$ and $Q$ have the opposite origins to those in the template one. In short, as shown in Fig.~\ref{fig1}(b), $Q$ is encoded from the features which need enhancement ($F_{q}$), and $K$, $V$ are encoded from the other ($F_{kv}$). We apply CA on each lines for $N$($N=3$) times, and use another CA on searching line to obtain the final output features: $F_{s, \text{CTA}}$.

We further create a novel structure called Registration-based Anatomical Attention Module (RAAM) to calculate an anatomical information mask $M_{A}$, whose transpose is taken as $M_{A}^{\text{T}}$ (see Fig.~\ref{fig1}(c)). As described above, the anatomical information is needed for lesion tracking. Thus, we create a matrix to provide the anatomical information for each of template and searching side. For template side, we assume all the voxels in $F_{t, reshape}$ are of the same importance, and we build a matrix $O \in \mathbb{R}^{L \times 1}$, in which all elements are 1. For searching side, we first use an affine registration method~\cite{marstal2016simpleelastix} to roughly align $I_{t}$ and $I_{s}$ by solving: $\mathcal{T}_{\text{Aff}} = \text{arg min} ||\mathcal{T}_{\text{Aff}}(I_{t}) - I_{s}||_{1}$. We choose to use an affine registration method instead a non-rigid one because the non-rigid registration is slow and provides restriction to the attention that limit the model’s ability to learn for local variation and details. Then, we can obtain a registration-based Gaussian map $G_{s} = \mathcal{T}_{\text{Aff}}(G_{t})$. Afterwards, we downsample $G_{s}$ to the size of $F_{s, ori}$ and reshape it to $G_{s, reshape} \in \mathbb{R}^{ D_{s}H_{s}W_{s}\times 1}$, which is defined as the matrix of searching side. Finally, $M_{A}$ can be calculated by the following formula:
\begin{equation}
M_{A} = O \otimes G_{s, reshape}^{\text{T}}
\end{equation}
where $\otimes$ is matrix multiplication operation, and $M_{A} \in \mathbb{R}^{L \times D_{s}H_{s}W_{s}}$. So the attention we use in CAT at each head (see Fig.~\ref{fig1}(b)) can be define as following:
\begin{equation}
\text{Attention}_{i}(Q, K, V) = \text{softmax}(\frac{(QW_{i}^{Q})(KW_{i}^{K})^\text{T}}{\sqrt{d_{k}}} + M_{A})(VW_{i}^{V}),
\end{equation}
where $W_{i}^{Q}$, $W_{i}^{K}$, $W_{i}^{V}$ are parameter matrices, $d_{k}$ is the dimension of key, $i \in \{1,...,h\}$ is the index of head and $h$ is the number of heads in multiple head attention. 

\subsection{Center Predictor and Training Loss} \label{sec2.4}
Similar to the head of detection networks, our center predictor consists of a classification branch and a regression branch, where each branch is a multilayer perceptron (MLP). The classification head is to classify if a voxel from the output is inside of a lesion, and the regression head is to regress the exact center position. In detail, after inputting the features $F_{s, \text{CTA}}$, the predictor outputs the classification results $\hat{Y}  \in \mathbb{R}^{1\times D_{s} H_{s}W_{s}} $ and center coordinates $\hat{C}\in \mathbb{R}^{3\times D_{s}H_{s}W_{s}}$. During training, we define the ground truth as a Gaussian map generated by Eq.~\ref{eq1} with the target center $c_{s}$ and the corresponding radius $r_{s}$. We downsample it to obtain the Gaussian label $G_{L}$ which matches the size of $\hat{Y}$, and obtain label $Y = \frac{G_{L} - min(G_{L})}{max(G_{L}) - min(G_{L})}$. L1 loss is used as the regression loss, which is formulated as: 
\begin{equation}
L_{r} = ||\hat{c} - c_{s}||_{1}, \ \hat{c} = \sum{\text{softmax}(\hat{Y}) * \hat{C}}
\end{equation}
where $\hat{c}$ is the final output of the center predictor, which we define as global regression. Meanwhile, a focal loss~\cite{cai2021deep} is used as the classification loss for auxiliary training:

\begin{equation}
L_{c} = \sum_{i} \begin{cases}
(1 - \hat{y}_{i})^{\alpha}\text{log}(\hat{y}_{i})& \text{if} \ y_{i}=1\\
(1 - y_{i})^{\beta}(\hat{y}_{i})^{\alpha}\text{log}(1 - \hat{y}_{i})& \text{otherwise}
\end{cases}
\end{equation}
where $y_{i}$ and $\hat{y_{i}}$ are the $i$-th elements in $Y$ and $\hat{Y}$, respectively, and $\alpha = \beta = 2$. 

\begin{table}
\caption{Lesion tracking comparison on Deep Lesion Tracking testing dataset. $^{*}$ represents the p value of paired t-test is smaller than 0.05.}\label{tab1}
\resizebox{\textwidth}{!}{
\begin{tabular}{l|c|c|c|c|c|c}
\hline
Method & \tabincell{c}{CPM$@$ \\ 10$mm$} & \tabincell{c}{CPM$@$ \\ Radius} \ & \tabincell{c}{MED$_{X}$ \\ ($mm$)} \ & \tabincell{c}{MED$_{Y}$ \\ ($mm$)} \ & \tabincell{c}{MED$_{Z}$ \\ ($mm$)} \ & \tabincell{c}{MED \\ ($mm$)} \ \\
\hline
Affine~\cite{marstal2016simpleelastix} & $48.33$ & $65.21$ & $4.1\pm5.0$ & $5.4\pm5.6$ & $7.1\pm8.3$ & $11.2\pm9.9$ \\
VoxelMorph~\cite{balakrishnan2018unsupervised} & $49.90$ & $65.59$ & $4.6\pm6.7$ & $5.2\pm7.9$ & $6.6\pm6.2$ & $10.9\pm10.9$ \\
LENS-LesioGraph~\cite{yan2020learning, yan2018deep} & $63.85$ & $80.42$ & $\mathbf{2.6\pm4.6}$ & $2.7\pm4.5$ & $6.0\pm8.6$ & $8.0\pm10.1$ \\
VULD-LesionGraph~\cite{cai2020deep, yan2018deep} & $64.69$ & $76.56$ & $3.5\pm5.2$ & $4.1\pm5.8$ & $6.1\pm8.8$ & $9.3\pm10.9$ \\
VULD-LesaNet~\cite{cai2020deep, yan2019holistic} & $65.00$ & $77.81$ & $3.5\pm5.3$ & $4.0\pm5.7$ & $6.0\pm8.7$ & $9.1\pm10.8$ \\
SiamRPN++~\cite{li2019evolution} & $68.85$ & $80.31$ & $3.8\pm4.8$ & $3.8\pm4.8$ & $4.8\pm7.5$ & $8.3\pm9.2$ \\
LENS-LesaNet~\cite{yan2020learning, yan2019holistic} & $70.00$ & $84.58$ & $2.7\pm4.8$ & $\mathbf{2.6\pm4.7}$ & $5.7\pm8.6$ & $7.8\pm10.3$ \\
DEEDS~\cite{heinrich2013mrf} & $71.88$ & $85.52$ & $2.8\pm3.7$ & $3.1\pm4.1$ & $5.0\pm6.8$ & $7.4\pm8.1$ \\
DLT-Mix~\cite{cai2021deep} & $78.65$ & $88.75$ & $3.1\pm4.4$ & $3.1\pm4.5$ & $4.2\pm7.6$ & $7.1\pm9.2$ \\
DLT~\cite{cai2021deep} & $78.85$ & $86.88$ & $3.5\pm5.6$ & $2.9\pm4.9$ & $4.0\pm6.1$ & $7.0\pm8.9$ \\
TransT~\cite{chen2021transformer} & $79.59$ & $88.99$ & $3.4\pm5.9$ & $5.4\pm6.1$ & $1.8\pm2.2$ & $7.6\pm7.9$ \\
\hline
TLT & $\mathbf{87.37}^{*}$ & $\mathbf{95.32}^{*}$ & $3.0\pm6.2$ & $3.7\pm5.2$ & $\mathbf{1.7\pm2.1}$ & $\mathbf{6.0\pm7.7}^{*}$ \\
\hline
\end{tabular}
}
\end{table}

\section{Experiments and Experimental Results}
\subsection{Dataset and Experiment Setup} \label{sec3.1} 
\textbf{Dataset.} We validate our method on a public dataset, DLS~\cite{cai2021deep}, which consists of CT image pairs inherited from DeepLesion~\cite{yan2018deeplesion}. There are 3008, 403 and 480 lesion pairs for training, validation,and testing in this dataset, respectively. Since the ground truth lesion center of all lesions in this dataset and the corresponding radius are defined, we could mutually track within a lesion pair. Therefore, a total of 906 and 960 directed lesion pairs are used for evaluation in validation and testing sets, respectively.

\noindent\textbf{Evaluation Metrics.} The center point matching (CPM) accuracy is selected to evaluate the performance of lesion matching. As defined in~\cite{cai2021deep}, a match will be counted correct when the Euclidean distance between ground truth and predicted centers is smaller than a threshold (@10$mm$: $min$(10$mm$, $r_{s}$), @Radius: $r_{s}$). The mean Euclidean distacne (MED) in $mm$ +/- standard deviation between ground truth and predicted centers, and its projections in each direction (denoted as $\text{MED}_X$, $\text{MED}_Y$ and $\text{MED}_Z$, respectively)~\cite{cai2021deep} are also used for model evaluation.

\noindent\textbf{Implementation Details.} The proposed method is implemented using PyTorch (v1.5.1). The network is optimized by Adam with initial learning rate of 0.0001 and trained for 300 epochs. The batch size is 4 and the number of parameters of the model is 5.98M. All CT volumes have been resampled to the isotropic resolution of $1mm$ before feeding into the network. This training setting is used in all deep learning-based methods selected for comparison. For the affine registration method~\cite{marstal2016simpleelastix} and DEEDS~\cite{heinrich2013mrf}, following the setting of~\cite{cai2021deep} and~\cite{heinrich2013mrf}, all CT volumes are resampled to a isotropic resolution of $2mm$.

\begin{table}
\caption{Ablation study on each module and different thresholds. $^{*}$ represents the p value of paired t-test is smaller than 0.05.}\label{tab2}
{
\resizebox{\textwidth}{!}{
\begin{tabular}{ccc|c|c|c|c|c|c}
\hline
\multirow{3}{*}{\shortstack{SSS}} & \multirow{3}{*}{\shortstack{RAAM}} & \multirow{3}{*}{\shortstack{Global\\Regressor}} & \multirow{3}{*}{\shortstack{CPM$@$\\10$mm$}} & \multicolumn{1}{c|}{\multirow{3}{*}{\shortstack{MED\\($mm$)}}} &  & Threshold & CPM$@$10$mm$ & MED($mm$) \\ \cline{7-9} 
&   &   &   & \multicolumn{1}{c|}{}   &  & 0.9   &  $83.57$   &  $6.80\pm8.12$    \\
&   &   &   & \multicolumn{1}{c|}{}  &  & 0.8   &  $83.99$   &   $6.65\pm7.99$  \\ \cline{1-5}
&   &   & $79.59$  & $7.58\pm7.91$ &  & 0.7 &  $\mathbf{87.37}^{*}$  &  $\mathbf{5.98\pm7.68}^{*}$   \\
\Checkmark &  &  & $84.78$ & $6.76\pm7.86$ &  & 0.6 &  $86.70$   &  $6.26\pm7.88$   \\
\Checkmark & \Checkmark & & 86.58 & $6.30\pm7.79$ &  & 0.5 &  $86.37$   &  $6.20\pm7.83$   \\
\Checkmark & \Checkmark & \Checkmark & $\mathbf{87.37}^{*}$ & $\mathbf{5.98\pm7.68}^{*}$ &  & 0.4  &  $85.01$  &  $6.39\pm7.98$   \\ \hline
\end{tabular}
}}
\end{table}

\subsection{Experimental Results and Discussion}     
\noindent\textbf{Model Comparison.} We took TransT~\cite{chen2021transformer} as baseline, and selected DLT and other state-of-the-art approaches in~\cite{cai2021deep} for comparison. Table \ref{tab1} shows the quantitative results of these methods. Our method yields a CPM$@10mm$ of 87.37, a CPM$@Radius$ of 95.32, and a MED of $6.0\pm7.7$, which outperforms all the compared methods in terms of both CPM and MED metrics. A paired t-test is used on CPM@$10mm$, CPM@$Radius$ and MED to perform statistical tests. Moreover, we observe that transformer-based methods, TransT and our TLT, both achieving large improvements in terms of MED$_{Z}$ compared with methods that use convolution to compute similarity. This may be because the Transformer focuses more on the information in the z-axis direction, which is also consistent with physician cognition.

\noindent\textbf{Ablation Study.} To evaluate the effectiveness of various configurations in our proposed method, we conduct ablation experiments from two aspects: module setting and threshold setting. A paired t-test is also used for statistical tests. Table~\ref{tab2} shows the experimental results. The results show that accuracy drops with each module change, which validates the competence of our proposed method. Meanwhile, it is observed that the threshold of 0.7 is much better than that of other thresholds. Therefore, we choose 0.7 as the thresholds $Tr$ in our TLT.

\noindent\textbf{Discussion.} As we observe, in ablation study, the SSS module leads to the biggest improvement. To verify this, we also perform ablation study with only one single module removed, as shown in Table 1 in supplementary materials. This happens when there are many small lesions in the dataset, such as lung nodules. If these small lesions are cropped on the original image, due to downsampling, the feature map will become very small, and in the last several downsampling processes will always become one voxel, which could lead to a decline in performance. The SSS solves this problem by selecting voxels on the last feature map. Even if only one voxel on the feature map is selected, this voxel can still obtain more surrounding information in the networks than without SSS. Meanwhile, based on our observations, we found that when the registration method failed, sometimes our model would fail as well. This is because we use registration to feed anatomical information to the transformer, and anatomical information helps the transformer accelerate convergence, which forms a dependency. In addition, when there are similar lesions in similar locations, such as two solid nodules at the edge of the right upper lung, and only a few layers difference in the z-axis direction, the model will also be confused. 

\section{Conclusion}
This paper presents a novel Transformer-based framework for lesion tracking by leveraging both the anatomical prior and the cross image relevance. We further introduce a global regression to integrate multi-scale information while using sparse selection strategy to reduce memory consumption. TLT achieves the state-of-the-art performance on DLT dataset, significantly exceeding previous methods in lesion tacking accuracy. Future work includes multi-institutional validation and reader studies to examine the efficiency improvement for physicians in clinical setting.
\indent \\

\noindent\textbf{Acknowledgment.} This work was funded by Science and Technology Innovation 2030-New Generation Artificial Intelligence Major Project (2021ZD0111104).

%
%
%
\bibliographystyle{splncs04.bst}
\bibliography{ref.bib}
\end{document}


\title{Supplementary Materials}
\titlerunning{Supplementary Materials}
%
\author{Wen Tang\inst{1}, Han Kang\inst{1}, Haoyue Zhang\inst{2}, Pengxin Yu\inst{1}, Corey Arnold\inst{2}, 
Rongguo Zhang \inst{1}\textsuperscript{(\Letter)}}
\authorrunning{Wen Tang et al.}
%
\institute{InferVision Medical Technology Co., Ltd.\\
\email{zrongguo@infervision.com}, \\
University of California, Los Angeles}
\maketitle              

\begin{table}
\centering
\caption{Ablation study with one single module removed}\label{tab2}
{
\begin{tabular}{ccc|c|c}
\hline
\multirow{2}{*}{\shortstack{SSS}} & \multirow{2}{*}{\shortstack{RAAM}} & \multirow{2}{*}{\shortstack{Global\\Regressor}} & \multirow{2}{*}{\shortstack{CPM$@$\\10$mm$}} & \multicolumn{1}{c}{\multirow{2}{*}{\shortstack{MED\\($mm$)}}} \\
&   &   &   & \multicolumn{1}{c}{}\\ \cline{1-5}
 & \Checkmark & \Checkmark & $84.02$ & $6.35\pm7.95$  \\
\Checkmark &  & \Checkmark & $85.11$ & $6.31\pm7.82$ \\
\Checkmark & \Checkmark & & $86.58$ & $6.30\pm7.79$  \\
\Checkmark & \Checkmark & \Checkmark & $\mathbf{87.37}^{*}$ & $\mathbf{5.98\pm7.68}^{*}$ \\ \hline
\end{tabular}
}
\end{table}

\begin{table}
\centering
\caption{Detailed stucture of our proposed method: TLT.}\label{tab1}

\begin{tabular}{cc}
\hline
\multicolumn{1}{c|}{template image} & searching image \\
\hline
\multicolumn{2}{c}{conv $7 \times 7 \times 7$, channel 32, stride 1.} \\
\hline
\multicolumn{2}{c}{conv $3 \times 3 \times 3$, channel 32, stride 1.} \\
\hline
\multicolumn{2}{c}{\multirow{3}{*}{
$\text{ResBlock} : \left[
\begin{aligned}
\text{conv} \ 3 \times 3 \times 3, \text{channel} \ 64 \\
\text{conv} \ 3 \times 3 \times 3, \text{channel} \ 64
\end{aligned}
\right] \times 2 $}}  \\
\multicolumn{2}{c}{}  \\
\multicolumn{2}{c}{} \\
\hline
\multicolumn{2}{c}{\multirow{3}{*}{
$\text{ResBlock} : \left[
\begin{aligned}
\text{conv} \ 3 \times 3 \times 3, \text{channel} \ 128 \\
\text{conv} \ 3 \times 3 \times 3, \text{channel} \ 128
\end{aligned}
\right] \times 2 $} }  \\
\multicolumn{2}{c}{} \\
\multicolumn{2}{c}{} \\
\hline
\multicolumn{2}{c}{\multirow{3}{*}{
$\text{ResBlock} : \left[
\begin{aligned}
\text{conv} \ 3 \times 3 \times 3, \text{channel} \ 192 \\
\text{conv} \ 3 \times 3 \times 3, \text{channel} \ 192
\end{aligned}
\right] \times 2 $} }  \\
\multicolumn{2}{c}{}\\
\multicolumn{2}{c}{}\\
\hline
\multicolumn{1}{c|}{SSS} &  \\
\hline
\multicolumn{2}{c}{reshape} \\
\hline
\multicolumn{1}{c|}{\multirow{4}{*}{
$\text{CA} : \left[
\begin{aligned}
\text{channel} \ 192, \text{head} \ 8\\
\text{Q:template features} \\ 
\text{K,V:searching features}
\end{aligned}
\right] \times 3 $} } &  \multirow{4}{*}{
$\text{CA} : \left[
\begin{aligned}
\text{channel} \ 192, \text{head} \ 8\\
\text{Q:searching features} \\ 
\text{K,V:template features}
\end{aligned}
\right] \times 3 $} \\
\multicolumn{1}{c|}{} & \\
\multicolumn{1}{c|}{} & \\
\multicolumn{1}{c|}{} & \\
\hline
\multicolumn{2}{c}{\multirow{4}{*}{
$\text{CA} : \left[
\begin{aligned}
\text{channel} \ 192, \text{head} \ 8\\
\text{Q:searching features} \\ 
\text{K,V:template features}
\end{aligned}
\right] \times 1 $} } \\
\multicolumn{2}{c}{}\\
\multicolumn{2}{c}{}\\
\multicolumn{2}{c}{}\\
\hline
\multicolumn{2}{c}{MLP: classification \& regression} \\
\hline
\end{tabular}
\end{table}

\begin{figure}
\includegraphics[width=\textwidth]{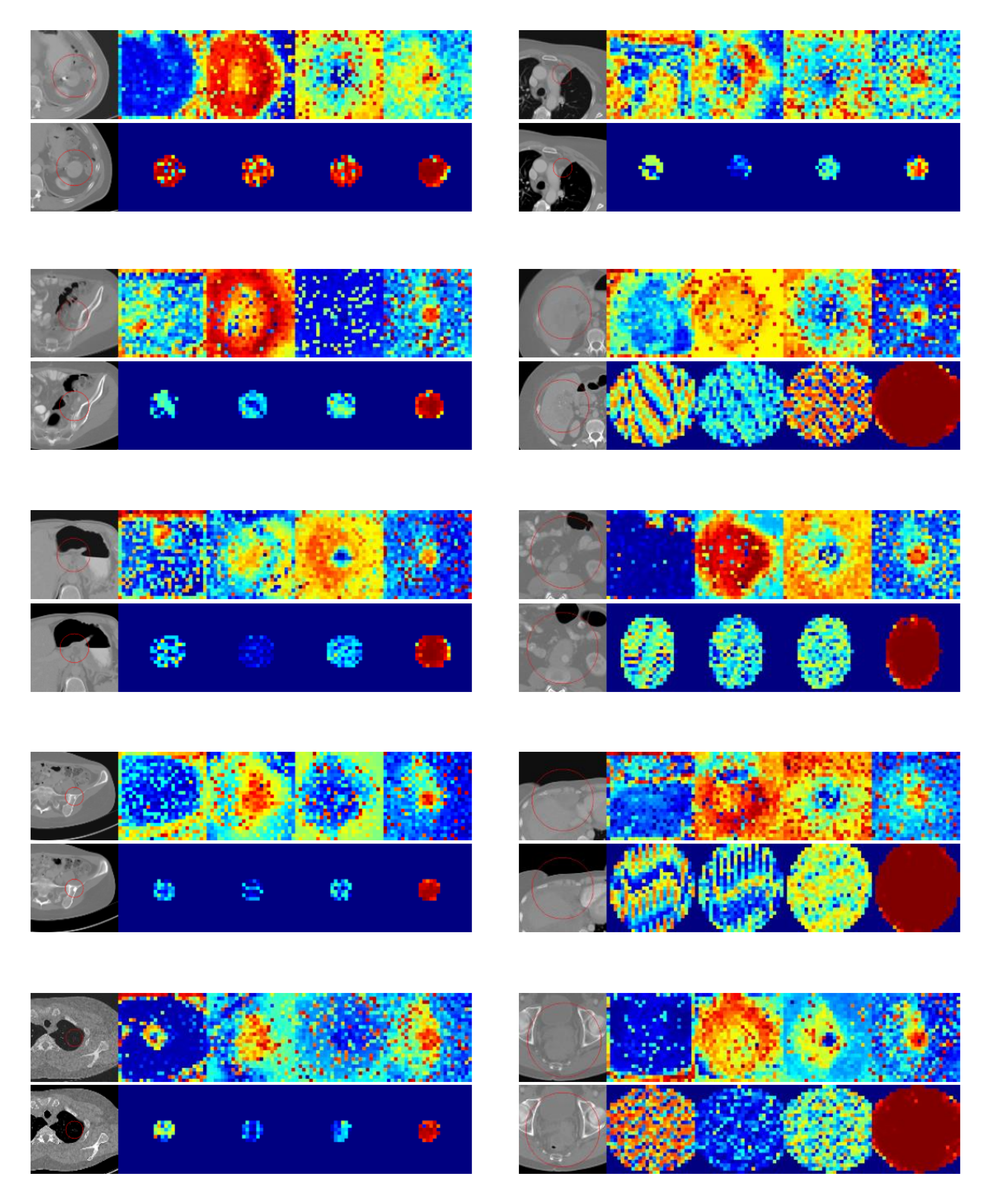}
\caption{Visualization of the attention maps for some representative pairs. The images above are from searching lines and the images below are from template lines. From left to right: the original input, cross attention map of each level (1-3), the final cross attention map before output. } \label{fig1}
\end{figure}